\documentclass[runningheads]{llncs}

\usepackage{eccv}

\usepackage{eccvabbrv}

\usepackage{graphicx}
\usepackage{booktabs}
\usepackage{bm}

\usepackage[accsupp]{axessibility}  

\usepackage{graphicx}
\usepackage{amsmath}
\usepackage{amssymb}
\usepackage{booktabs}
\usepackage{multirow}
\usepackage{threeparttable}
\usepackage{color}
\usepackage{enumitem}
\usepackage{pifont}
\usepackage{arydshln}
\usepackage{colortbl}
\usepackage{wrapfig}
\usepackage[noend]{algpseudocode}
\usepackage{algorithmicx,algorithm}
\definecolor{bblue}{rgb}{0,150,230}
\definecolor{mygray}{gray}{.9}

\newcolumntype{I}{!{\vrule width 1pt}}

\definecolor{rui}{RGB}{123,104,238}
\definecolor{ggray}{RGB}{127,127,127}

\newcommand{\makesupptitle}[1]{
	\begin{center}
		{\Large \bf #1 \par}
		{
		\large
		\lineskip .5em
		\par
		}
		\vskip .5em
	\end{center}
	
}
\newcommand\blfootnote[1]{%
\begingroup
\renewcommand\thefootnote{}\footnote{#1}%
\addtocounter{footnote}{-1}%
\endgroup
}

\makeatletter
\newcommand{\thickhline}{%
    \noalign {\ifnum 0=`}\fi \hrule height 1pt
    \futurelet \reserved@a \@xhline
}
\makeatother

\newcommand{\pub}[1]{\color{gray}{\tiny{[{#1}]}}}
\newcommand{\cmark}{\ding{52}}%
\usepackage{caption}
\usepackage[pagebackref,breaklinks,colorlinks,citecolor=eccvblue]{hyperref}

\usepackage{orcidlink}

\begin{document}

\title{Navigation Instruction Generation with BEV Perception and Large Language Models}
%

\titlerunning{\textsc{BEVInstructor}}

\author{Sheng Fan \and
Rui Liu \and
Wenguan Wang$^\dag$ \and
Yi Yang \\
}

\authorrunning{S.~Fan et al.}

\institute{ReLER, CCAI, Zhejiang University\\
\url{https://github.com/FanScy/BEVInstructor}}

\maketitle

\begin{abstract}
	Navigation instruction generation, which requires embodied agents to describe the navigation routes, has been of great interest in robotics and human-computer interaction. Existing studies directly map the sequence of 2D  perspective observations to route descriptions. Though straightforward, they overlook the geometric information and object semantics of the 3D environment. To address these challenges, we propose \textsc{BEVInstructor}, which incorporates Bird’s Eye View (BEV) features into Multi-Modal Large Language Models (MLLMs) for instruction generation. Specifically, \textsc{BEVInstructor} constructs a Perspective-BEV Visual Encoder for the comprehension of 3D environments through fusing BEV and perspective features. To leverage the powerful language capabilities of MLLMs, the fused representations are used as visual prom-pts for MLLMs, and perspective-BEV prompt tuning is proposed for parameter-efficient updating. Based on the perspective-BEV prompts, \textsc{BEVInstructor} further adopts an instance-guided iterative refinement pipeline, which improves the instructions in a progressive manner. \textsc{BEVInstructor} achieves impressive performance across diverse datasets (\ie, R2R, REVERIE, and UrbanWalk). 
\end{abstract}

\blfootnote{$^\dag$~Corresponding author: \textit{Wenguan Wang}.}

\section{Introduction}
\label{sec:intro}

\begin{figure}[t]
	\begin{center}
		\includegraphics[width=0.98\linewidth]{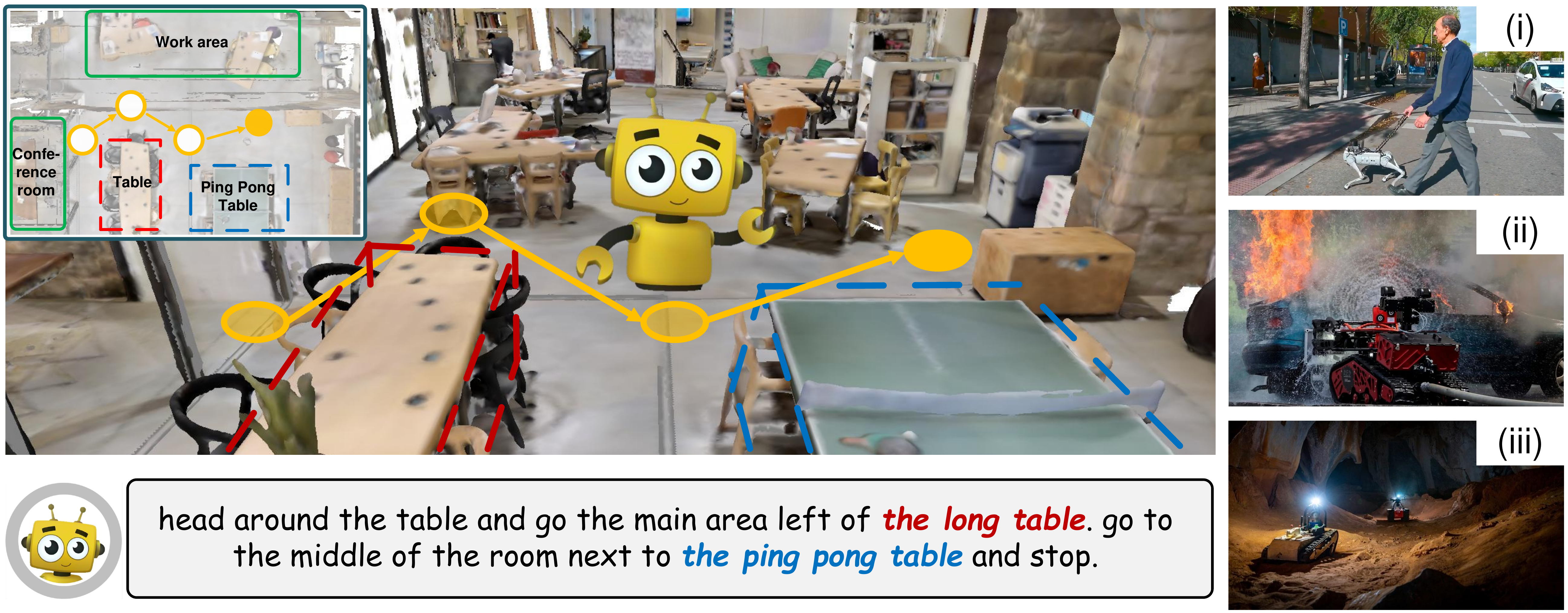}
	\end{center}
	\captionsetup{font=small}
	\caption{\small{\textsc{BEVInstructor}$_{\!}$ verbalizes$_{\!}$ concise$_{\!}$ navigation$_{\!}$ instructions. Navigation instruction generation is of great value to a wide range of tasks, (\romannumeral1) assisting in navigation for blind individuals, (\romannumeral2) executing long-term tasks with automatic progress reporting, and (\romannumeral3) conducting autonomous search and rescue operations in disaster areas.}}
	\label{fig:intro}
\end{figure}

Navigation instruction generation, serving as a crucial interface between intelligent robotics and human interaction, has garnered significant attention in various fields, including robotics~\cite{goeddel2012dart}, psychology~\cite{vanetti1988communicating}, and cognitive science~\cite{evans1981environmental, kuipers1978modeling}. This research aims to describe the navigation route precisely based on the observations. The process involves analyzing a series of visual inputs and subsequently converting them into natural language instructions. The generated instructions are required to incorporate the details for accurate replication of the navigated path. Navigation instruction generation plays a crucial role in fostering trust between humans and machines. It provides intuitive feedback to humans and guides them to accomplish goals, such as aiding visually impaired individuals~\cite{huang2022assister} and explaining the agent's plan~\cite{wang2023lana,wang2023learning} (see Fig.\!~\ref{fig:intro}). Furthermore, embodied agents are expected to communicate with humans for efficient collaboration, instead of executing instructions only~\cite{huang2022assister,wang2022counterfactual,wang2023lana,wang2020active,wang2021structured,wang2022towards,wang2023dreamwalker,chen2023omnidirectional,an2023etpnav,wang2023active}. 

Early solutions~\cite{dale2004using,look2005location,goeddel2012dart} in instruction generation utilize hand-crafted rules or templates that fill a predetermined format with specific details. While straightforward, these approaches lack flexibility. Subsequent studies employ neural networks to facilitate end-to-end learning for instruction generation, \eg, LSTM~\cite{fried2018speaker, tan2019learning,wang2022counterfactual,zeng2023kefa} or Transformer~\cite{wang2023lana,wang2024pasts,wang2023learning}. Recent Multi-Modal Large Language Models (MLLMs) showcase immense capabilities of vision-language understanding and generation~\cite{openai2023gpt4,wang2024visionllm,zhu2023minigpt,li2023videochat,touvron2023llama,yang2024doraemongpt}. MLLMs take advantage of cross-modal transfer, allowing knowledge to be shared between multi-modal domains~\cite{yang2023octopus,huang2023voxposer,wu2023tidybot,driess2023palm,brohan2023rt}. Despite their promising performance on various vision-language tasks~\cite{openai2023gpt4,touvron2023llama}, they cannot fully satisfy the requirements of navigation instruction generation in a zero-shot manner (see Table~\ref{table:R2RLLMig}).
Specifically, MLLMs are pre-trained on extensive image-text pairs, primarily involving isolated images from a third-person view. In contrast, navigation instruction generation relies on a sequence of egocentric observations from an embodied agent~\cite{grauman2022ego4d}. 
This poses challenges for MLLMs in understanding spatial context from navigation trajectories, especially in complex 3D environments. More importantly, such embodied (egocentric) perception requires a comprehensive scene understanding of the 3D physical world, interpreting objects and actions to generate detailed instructions. However, existing studies~\cite{wang2022counterfactual,wang2023lana,fried2018speaker,tan2019learning,wang2023learning,zheng2019reasoning} often rely on 2D perspective features as visual representations, ignoring the 3D scene geometry and object semantics~\cite{li2023delving,ma2022vision}. This underscores the need for more advanced solutions capable of integrating 3D spatial understanding to improve the accuracy and relevance of navigation instructions.

As a response, we propose \textsc{BEVInstructor}, an iterative instruction generator driven by BEV perception and MLLMs. \textsc{BEVInstructor} develops a \textit{BEV encoder} to reconstruct 3D information from perspective features under the supervision of 3D detection. This allows to preserve 3D geometry and object semantics of the environment. The encoded BEV features are combined with the perspective features, thereby enriching the visual representations. Then \textsc{BEVInstructor} enhances the capability of the MLLMs by finetuning on navigation instruction-specific data through a \textit{parameter-efficient update} strategy. In addition, \textit{iterative refinement} is proposed to progressively enhance instruction generation, leveraging the powerful language capabilities of MLLMs.

Specifically, \textsc{BEVInstructor} processes a sequence of embodied observations. It adopts a \textit{BEV encoder} to aggregate the multi-view image features into the BEV grid features through a 2D-3D view transformation (\S\ref{sec:visualencoder}). Then it uses a \textit{Perspective-BEV fusion} module to fuse the BEV features with perspective features, converting the fused embeddings into shorter tokens to prevent excessively long inputs for the MLLM. We also devise perspective-BEV prompt tuning for parameter-efficient updating (\S\ref{sec:alignment}), with trainable parameters constituting only \textbf{7.2}\% of the entire framework. Prior research in cognitive science~\cite{lynch1964image} has validated the significance of key instances and landmarks in human route description. This motivates us to propose \textit{instance-guided iterative refinement} (\S\ref{sec:refinement}). Initially, \textsc{BEVInstructor} identifies the key instances and generates the corresponding landmark tokens along the path, leveraging the rich object semantics encoded in the perspective-BEV embeddings. Subsequently, it organizes the complete instructions conditioned on these landmark drafts. After multi-turn refinement, \textsc{BEVInstructor} produces high-quality instructions that include more concise details about the 3D environment.

We conduct extensive experiments on indoor R2R~\cite{anderson2018vision}, REVERIE~\cite{reverie} and outdoor UrbanWalk~\cite{huang2022assister} datasets. Compared with the state-of-the-art navigation instruction algorithms~\cite{huang2022assister,wang2022counterfactual,wang2023lana}, our \textsc{BEVInstructor} attains better performances across all datasets. Especially, \textsc{BEVInstructor} achieves \textbf{12.6}\% and \textbf{8.3}\% CIDEr gains on REVERIE val seen and unseen splits, respectively, compared with previous best methods. This suggests that the BEV features effectively integrate the 3D scene information into the MLLM, thereby establishing a connection between real-world perceptions and human languages.

\section{Related Work}
\label{sec:related}

\noindent\textbf{Navigation Instruction Generation.} The study of navigation instruction generation can date back to the 1960s~\cite{lynch1964image}. Instruction generation~\cite{curry2015generating} has garnered significant interest across various fields, \eg,  robotics~\cite{goeddel2012dart}, psychology~\cite{vanetti1988communicating}, and cognitive science~\cite{evans1981environmental, kuipers1978modeling} for a long time, yet attained far less attention in embodied vision. Early studies predominantly rely on hand-crafted rules~\cite{dale2004using} or templates~\cite{look2005location,goeddel2012dart}, binding the format of generated instructions. While these methods are effective in producing high-quality sentences tailored to specific scenes, they demand significant linguistic expertise and extensive manual effort. To alleviate this inflexibility, several studies employ neural network~\cite{fried2018speaker,fried2017unified} to facilitate end-to-end learning. Subsequent efforts~\cite{fried2018speaker, tan2019learning,wang2022counterfactual} utilize LSTM-based speakers integrated with instruction-following agents, enabling simultaneous training on pairs of path navigation and instructions navigation. They leverage the sequential processing strengths of LSTMs to better understand and generate navigation instructions that accurately reflect the temporal process of navigating. Additionally, motivated by the success of Transformer~\cite{vaswani2017attention} in the natural language processing domain, a new wave of research~\cite{wang2023lana,wang2024pasts,wang2023learning} has emerged to leverage the advanced capabilities of Transformer to further improve generation performance. Existing efforts delve into understanding the foundational principles of how humans construct route descriptions~\cite{ward1986turn,allen1997knowledge,lovelace1999elements} and explore the qualities that make instructions easy to follow~\cite{look2005location,waller2007landmarks,richter2008simplest}. These studies emphasize that crucial landmarks and concise topological descriptions play a crucial role in the description of wayfinding. In light of these, recent studies~\cite{agarwal2019visual,moudgil2021soat,wang2022less,cui2023grounded} lean on the process of landmark grounding to improve the instructions.

From the perspective of network architectures, a text encoder with powerful representation capacity significantly enhances output quality. However, there is a lack of dedicated research on the application of MLLMs for creating navigation instructions. In this work, we explore the potential of incorporating MLLMs endowed with superior linguistics for navigation instruction generation. Furthermore, \textsc{BEVInstructor} is designed to initially identify critical landmarks as drafts, aiding in forming comprehensive instructions. We devise an instance-guided iterative refinement process, which decomposes the generation into two stages. This allows for iterative refinement and enrichment of the instructions.

\noindent\textbf{Scene Understanding.} Scene understanding has emerged as a pivotal aspect of perception, navigation, and interaction with humans and environments~\cite{baruch2021arkitscenes,dai2017scannet,li2023delving}. Traditional SLAM systems~\cite{durrant2006simultaneous} leverage data from different sensors, such as LiDAR and cameras, to build maps. They facilitate the robot to perceive depth and structure but exhibit limitations in comprehending the scene semantics. Several efforts develop semantic spatial representations~\cite{cartillier2021semantic,henriques2018mapnet,an2023bevbert,liu2023bird,chen2022weakly,hong2023learning,wang2023gridmm,liu2024vol} or neural scene representations~\cite{mildenhall2021nerf,kwon2023renderable}, showing effectiveness across diverse scenes. Recently, BEV perception~\cite{li2022bevformer,philion2020lift,li2023fbbev,CaDDN} is proposed to infer the 3D geometry by projecting the multi-view images onto the BEV plane.

Current research~\cite{wang2022counterfactual,wang2023lana,fried2018speaker,tan2019learning,wang2023learning} in instruction generation relies on perspective features. While this provides a foundational understanding of the environment, it ignores critical aspects like scene geometry and object semantics, often resulting in suboptimal performance in complex environments~\cite{tan2019learning,wang2022counterfactual}. Our approach aims to enhance the instruction generation process by integrating both perspective and BEV features. This fusion achieves a more holistic understanding of the scene, thereby facilitating the generation of higher-quality instructions.

\noindent\textbf{Multi-Modal Large Language Models (MLLMs).} MLLMs have surged in popularity and application. Although primarily trained on text data, initial studies~\cite{liu2024visual, zhu2023minigpt, li2023blip} have demonstrated that pre-trained MLLMs can process visual information by fine-tuning the vision encoder via a learnable interface.
The profound impact of this simple yet efficient approach drives advancements in MLLMs. Existing open-source MLLMs can be broadly classified into three categories based on the approach of vision fusion: query-based, projection-based and parameter-efficient tuning-based. Motivated by the success of BLIP-2~\cite{li2023blip}, numerous efforts~\cite{zhang2023video,   li2023videochat, zhu2023minigpt, dai2023instructblip, wang2024visionllm} investigate the use of a lightweight Q-Former to efficiently extract vision information. Albeit simple, ~\cite{li2024llava,yin2024lamm, maaz2023video, liu2024visual, pi2023detgpt, su2023pandagpt} adopt the linear layers to project the vision embeddings into the language embedding space. Several solutions~\cite{luo2023cheap, zhang2023llama,gao2023llamaadapterv2} train on multi-modal data simultaneously by parameter-efficient tuning, \ie, LoRA~\cite{hu2022lora} and Adapter~\cite{houlsby2019parameter, sung2022vl}.

Though impressive, these methods only address the alignment of individual image-text pairs, neglecting the interaction between the egocentric observation and 3D spatial scene understanding in the task of navigation instruction generation, \ie, fine-grained vision signals about landmarks and objects. 
Consequently, our research also delves into the effectiveness of scene understanding in improving the generation of navigation instructions by MLLMs.

\section{Methodology}
\label{sec:methodology}

\subsection{Overview}

\noindent\textbf{Problem Definition.} The goal of the instruction generation task is to verbalize the navigation trajectory of an embodied agent using natural language. Here we formulate the task under R2R~\cite{anderson2018vision} setting. The agent observes a navigation path and collects a sequence of perspective observations $\mathcal{O}=\{O_t\}_{t=1}^{T}$ along with actions $\mathcal{A}=\{a_t\}_{t=1}^{T}$. Each observation $O_t$ contains $K$ multi-view images of its surroundings with the orientation angles $\{\delta_{t,k}\}_{k=1}^K$. These RGB images are encoded as $\{\bm{F}_{t,k}\!\in\!\mathbb{R}^{{D_p}\times{HW}}\}_{k=1}^K$. $H$ and $W$ are the spatial shape of image features, $D_p$ is the channel dimension. The action embedding $\bm{a}_t\in\!\mathbb{R}^D$ is represented by the corresponding feature $\bm{F}_{t,a}$ of the action view $\delta_{t,a}$ ($0\leq a\leq K$). Based on the observation-action sequence $\{O_t,a_t\}_{t=1}^{T}$, the agent is required to produce an instruction $\mathcal{X}\!=\!\{\bm{x}_l\!\in\!\mathbb{R}^{D}\}_{l=1}^{L}$ with $L$ words in an autoregressive style ($D$ is the embedding dimension and $\rm \Theta$ is the model parameters):

\begin{equation}
 \begin{aligned}
\max _{\rm \Theta} \sum\nolimits_{l=1}^{L} \log P_{\rm \Theta}(\bm{x}_l|\bm{x}_{<l}, \mathcal{O}, \mathcal{A}).
 \end{aligned}
 \label{eq:autogre}
 \end{equation}

 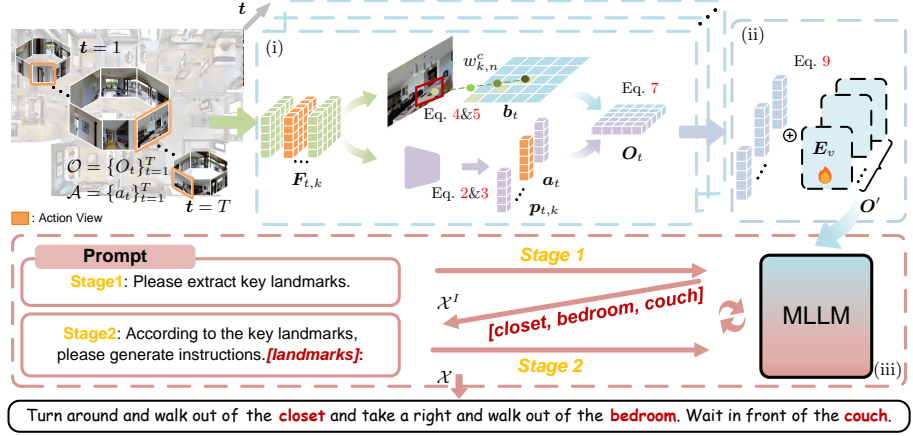
\begin{figure*}[t]
	  \centering
		  \includegraphics[width=0.99 \linewidth]{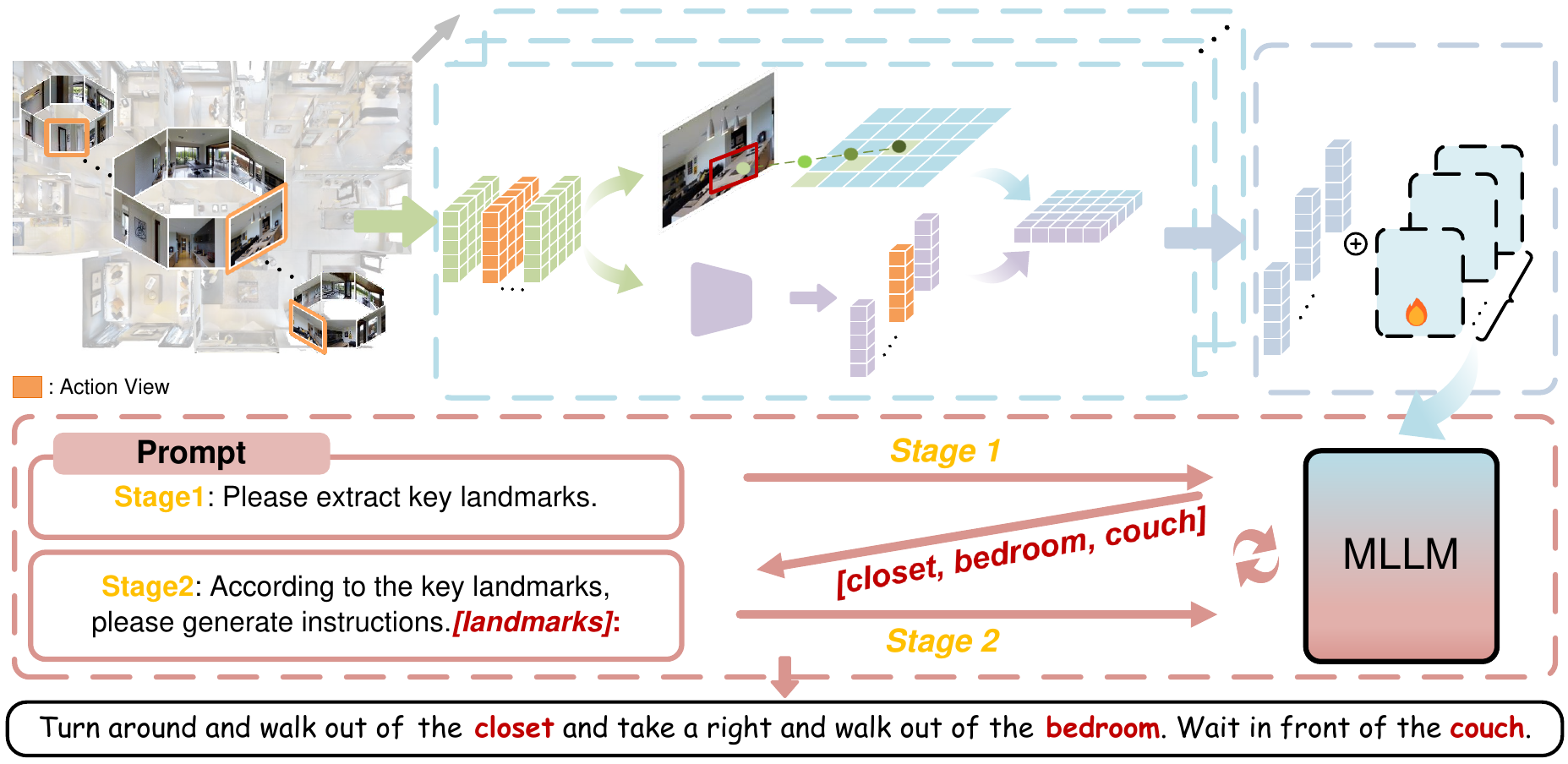}
		  \put(-182, 90){\scalebox{0.65}{Eq.~\ref{eq:panoemb}\&\ref{eq:actemb}}}
		  \put(-185, 120){\scalebox{0.65}{Eq.~\ref{eq:bevem}\&\ref{eq:bevweight}}}
		  \put(-110, 130){\scalebox{0.65}{Eq.~\ref{eq:bevprompt}}}
		  \put(-45, 140){\scalebox{0.65}{Eq.~\ref{eq:concatprompt}}}
		  \put(-315, 145){\scalebox{0.75}{$\bm{t}=1$}}
		  \put(-275, 85){\scalebox{0.75}{$\bm{t}=T$}}
		  \put(-255, 160){\scalebox{0.75}{$\bm{t}$}}
		  \put(-110, 105){\scalebox{0.75}{$\bm{O}_t$}}
		  \put(-38, 107){\scalebox{0.75}{$\bm{E}_v$}}
		  \put(-20, 85){\scalebox{0.75}{$\bm{O}'$}}
		  \put(-235, 95){\scalebox{0.75}{$\bm{F}_{t,k}$}}
		  \put(-155, 121){\scalebox{0.75}{$\bm{b}_{t}$}}
		  \put(-170, 140){\scalebox{0.75}{${w}^c_{k,n}$}}
		  \put(-145, 85){\scalebox{0.75}{$\bm{p}_{t,k}$}}
		  \put(-140, 95){\scalebox{0.75}{$\bm{a}_{t}$}}
		  \put(-320, 100){\scalebox{0.75}{$\mathcal{O}=\{O_t\}_{t=1}^{T}$}}
		  \put(-320, 90){\scalebox{0.75}{$\mathcal{A}=\{a_t\}_{t=1}^{T}$}}
		  \put(-180, 48){\scalebox{0.75}{$\mathcal{X}^I$}}
		  \put(-180, 20){\scalebox{0.75}{$\mathcal{X}$}}
		  \put(-245, 145){\scalebox{0.75}{(\romannumeral1)}}
		  \put(-65, 150){\scalebox{0.75}{(\romannumeral2)}}
		  \put(-15, 23){\scalebox{0.75}{(\romannumeral3)}}
	\caption{$_{\!}$Overview$_{\!}$ of$_{\!}$ \textsc{BEVInstructor}$_{\!}$ for$_{\!}$ navigation$_{\!}$ instruction$_{\!}$ generation. (\romannumeral1) BEV incorporates perspective embeddings by querying for 3D scene understanding  (\S\ref{sec:visualencoder}), (\romannumeral2) we adopt BEV-Perspective prompt tuning for the cross-modal alignment with MLLMs (\S\ref{sec:alignment}), (\romannumeral3) the instructions are generated and improved progressively through multiple refinements (\S\ref{sec:refinement}). Please refer to \S\ref{sec:methodology} for more details.}
	\label{fig:method}
\end{figure*}
 
\noindent\textbf{Core Idea.} We propose \textsc{BEVInstructor}, a new navigation instruction generator built upon LLaMA~\cite{touvron2023llama} (Fig.\!~\ref{fig:method}). To encode the semantic and geometric information of the 3D environment, BEV features are introduced and combined with 2D perspective features in \textit{Perspective-BEV Visual Encoder} (\S\ref{sec:visualencoder}). To exploit the capacity of cross-modal alignment in MLLMs, the visual embeddings are considered as visual prompts and fed into the \textit{Perspective-BEV Prompt Tuning} (\S\ref{sec:alignment}). Furthermore, we devise \textit{Instance-Guided Iterative Refinement} (\S\ref{sec:refinement}) to improve the quality of the generated instructions in a progressive manner.

\subsection{Perspective-BEV Visual Encoder}
\label{sec:visualencoder}
\noindent\textbf{Perspective Embedding.} The perspective embedding $\bm{P}_t=\{\bm{p}_{t,k}\in\mathbb{R}^D\}$ is built upon the multi-view features $\{\bm{F}_{t,k}\}_{k=1}^{K}$ of the surroundings with different view angles. To maintain the direction indication information, the orientation representation is incorporated into the perspective embedding of each view:
\begin{equation}
\begin{aligned}
  \bm{p}_{t, k}=\mathcal{E}^p(\bm{F}_{t,k})+\mathcal{E}^{\delta}(\delta_{t,k})+\bm{E}_t+\bm{E}_o \in \mathbb{R}^D,
\end{aligned}
\label{eq:panoemb}
\end{equation}
where $\mathcal{E}^p$ and $\mathcal{E}^{\delta}$ represent a linear layer, $\bm{E}_t\in \mathbb{R}^D$ and $\bm{E}_o\in \mathbb{R}^D$ denote the learnable embeddings of time step $t$ and observation token type, respectively. Analogously, the embedding of action (view) is formulated as:
\begin{equation}
\begin{aligned}
  \bm{a}_{t}=\mathcal{E}^a(\bm{F}_{t, a})+\mathcal{E}^{\delta}(\delta_{t,a})+\bm{E}_t+\bm{E}_a \in \mathbb{R}^D,
\end{aligned}
\label{eq:actemb}
\end{equation}
where $\mathcal{E}^a$ indicates a linear layer and $\bm{E}_a\in \mathbb{R}^D$ is the learnable embedding of action token type.

\noindent\textbf{BEV Embedding.}
\label{sec:bevemb}
Previous studies~\cite{wang2022counterfactual,wang2023lana,fried2018speaker,tan2019learning} adopt the 2D perspective features as visual representations of the observations $\mathcal{O}$ in the 3D environment. However, these 2D features only capture limited semantic information and geometry, easily leading to ambiguous path descriptions. To further enhance the visual representations for the 3D environment, BEV features are introduced to encode the spatial scene understanding. The BEV encoder assigns each BEV query $\bm{Q}(x,y)\in~\!\!\mathbb{R}^D$ located at $(x,y)$ on the BEV plane ($H_b\times W_b$) with a set of 3D reference points ${\mathcal{P}_k}(x,y,z_n)$. The BEV encoder projects them to sample the feature $\bm{F}_{t,k}$. Then the multi-view features are aggregated into the BEV grid features $\{\bm{b}_{t}(x,y)\in\mathbb{R}^{D_b}\}_{x=1,y=1}^{H_b,W_b}$ as (the subscripts $x,y$ are omitted for simplicity):
\begin{equation}
\begin{aligned}
\bm{b}_t=\sum\nolimits_{k=1}^{K}\sum\nolimits_{n=1}^{N_{\text{ref}}}{\mathcal{F}^d}\big(\bm{Q}(x,y),{\mathcal{P}_k}(x,y,z_n),\bm{F}_{t,k}\big)\cdot{{w}^c_{k,n}} \in~\mathbb{R}^{D_b} ,
\end{aligned}
\label{eq:bevem}
\end{equation}
where $\mathcal{F}^d$ is the BEV encoder with deformable attention layers~\cite{zhu2020deformable}. $\mathcal{F}^d$ uses $\bm{Q}(x,y)$ to sample the corresponding image feature $\bm{F}_{t,k}$ and $N_{\text{ref}}$ represents the number of 3D reference points. Since different reference points may be projected on the same image pixels to sample the feature, a depth consistency weight ${w}^c_{k,i}$ is introduced to distinguish them by predicting the weights of different depths:
\begin{equation}
	\begin{aligned}
	{w}^c_{k,n} = \frac{{\mathcal{E}^c}(\bm{F}_{t,k})\cdot{\mathcal{D}}({\mathcal{P}_k}(x,y,z_n))}{\big\|{\mathcal{E}^c}(\bm{F}_{t,k})\big\|\cdot \big\|{\mathcal{D}}({\mathcal{P}_k}(x,y,z_n))\big\|} \in[0,1],
	\label{eq:bevweight}
	\end{aligned}
	\end{equation}
	where $\mathcal{D}$ denotes a parameter-free operation that converts the 3D reference point ${\mathcal{P}_k}(x,y,z_n)$ into a depth distribution vector, and $\mathcal{E}^c$ is a depth network to predict the depth distribution vector of the projected image pixel based on $\bm{F}_{t,k}$. By calculating the cosine similarity between them, the depth consistency can guarantee the sampling quality in the 3D space\cite{li2023fbbev,li2022bevdepth}. The BEV encoder is trained under the supervision of 3D detection. The detection heads \cite{li2022bevformer,WangGZWZ021} with $\ell_1$ loss and cross-entropy loss are used to supervise 3D bounding box regression and semantic classification, respectively (more detailed in Appendix). Then the frozen BEV encoder $\mathcal{F}^d$ is adopted to sample the image features into the BEV plane as $\bm{B}_t=\{\bm{b}_{t}(x,y)\}_{x=1,y=1}^{H_b,W_b}$. 

\noindent\textbf{Perspective-BEV Fusion.} The perspective embedding $\bm{P}_t$ preserves rich visual cues in multi-view images. It is complementary to the BEV embedding $\bm{B}_t$ that mainly represents the 3D geometric information. Hence, \textit{perspective-BEV fusion} is proposed for comprehensive scene understanding. This module consists of several standard transformer layers $\mathcal{F}^o$ for attending to the spatial relationships between $\bm{B}_t$ and $[\bm{P}_t, \bm{a}_t]$ ($[,]$ denotes the \textit{concatenation} operation):
\begin{equation}
 \begin{aligned}
 \bm{O}_t = \mathcal{F}^o(\bm{B}_t,[\bm{P}_t, \bm{a}_t]) \in \mathbb{R}^{D\times {H_b}{W_b}}.
 \label{eq:fusion}
 \end{aligned}
\end{equation}

Through the \textit{perspective-BEV fusion} module, the fused embedding, \ie, the complete observation embedding $\bm{O}_t$, is served as the visual token and fed into MLLMs (Eq.~\ref{eq:autogre}) for instruction generation. However, given a large number of visual tokens (\eg $H_bW_b$ tokens for each step), aggregating such long tokens poses a significant challenge for MLLMs. In Table~\ref{table:diagnostic}, directly feeding all visual tokens into MLLMs results in excessive computational burden, and makes it difficult to capture critical semantic information. Therefore, we design a lightweight transformer $\mathcal{Q}$ with $N_q$ learnable queries to map $\bm{B}_t$ into a fixed number of tokens:
\begin{equation}
 \begin{aligned}
 \bm{O}_t = \mathcal{Q}(\mathcal{F}^o(\bm{B}_t,[\bm{P}_t, \bm{a}_t])) \in \mathbb{R}^{D\times N_q},
 \label{eq:bevprompt}
 \end{aligned}
\end{equation}
where $N_q$ is independent of the BEV dimension ($N_q\ll H_bW_b$). It reduces the number of visual tokens to $N_q$ and feeds the most useful tokens for instruction generation~\cite{li2023blip}. We conduct extensive experiments to confirm the effectiveness of our proposed perspective-BEV fusion (\S\ref{sec:experiments}). In Table~\ref{table:fusionablation}, we compare our fusion method with other approaches and find that our fusion method is notably more effective in boosting performance. 

\subsection{Perspective-BEV Prompt Tuning}
\label{sec:alignment}

The MLLMs~\cite{openai2023gpt4,touvron2023llama} typically utilize extensive corpora of vision-language pairs and project visual and linguistic information into a unified representation space. They encode numerous world knowledge acquired from massive data and possess strong capabilities in multi-modal tasks~\cite{maaz2023video, liu2024visual, zhu2023minigpt}. However, directly using general-purpose MLLMs for instruction generation fails to capture intricate details due to the complexity of scenarios (Table~\ref{table:R2RLLMig}). Additionally, the large size of MLLMs makes them costly to train from scratch. Thus, we propose perspective-BEV prompt tuning to exploit the scene geometry and unleash the cross-modal potential of MLLMs. Our perspective-BEV prompt tuning is parameter-efficient and incorporates 3D geometry into prompts. While our proposal is agnostic to the model, \ie, $P_{\rm \Theta}$ (Eq.~\ref{eq:autogre}) is a generic multi-modal learner, we formulate it with LLaMA~\cite{touvron2023llama} in a parameter-efficient updating manner:
\begin{equation}
 \begin{aligned}
\max _{\rm \Theta^*\cup\Psi} \sum\nolimits_{l=1}^{L} \log P_{\rm \Theta\cup\Psi}(\bm{x}_l|\bm{x}_{<l}, \bm{O}_{1:T}),
\label{eq:tuning}
 \end{aligned}
\end{equation}
where $\rm \Psi$ is the additional parameters for prompt tuning ($|\rm \Psi|\ll|\rm \Theta|$), $\rm \Theta^*$ is the finetuned parameters of $\rm \Theta$ ($|\rm \Theta^*|\ll|\rm \Theta|$), and $\bm{O}_{1:T}$ is the visual embedding sequence of $\mathcal{O}$ and $\mathcal{A}$ in Eq.~\ref{eq:autogre}.

\noindent\textbf{Perspective-BEV Prompt.} To overcomes the issues of catastrophic forgetting, $N_p$ learnable embeddings $\bm{E}_v\in \mathbb{R}^{D\times N_p}$ are inserted into the visual embedding $\bm{O}_{1:T}$ as perspective-BEV prompts $\bm{O}'$:
\begin{equation}
 \begin{aligned}
\bm{O}' = \bm{O}_{1:T} \oplus  \bm{E}_{v}\in \mathbb{R}^{D\times TN_p},
\label{eq:concatprompt}
 \end{aligned}
\end{equation}
where $\oplus$ indicates broadcast and addition on the sequence length dimension.
Then they are fed into the transformer layer with the text tokens $\bm{x}_{<l}\in \mathbb{R}^{D\times (l-1)}$:
\begin{equation}
 \begin{aligned}
[\bm{O}',\bm{x}_{<l}]_{m+1}=\mathcal{F}^{\text{LM}}_m([\bm{O}',\bm{x}_{<l}]_m),
\label{eq:llminput}
 \end{aligned}
\end{equation}
where $\mathcal{F}^{\text{LM}}_m$ is the $m$-th transformer layer in LLaMA.

\noindent\textbf{Parameter-Efficient Updating.} To stabilize the training process and modulate the deep features, we modify the vanilla attention mechanism with the self-attention and linear layers at the last $N_a$ layers of LLaMA. Specifically, for the self-attention part, zero-initialized attention~\cite{zhang2023llama} is adopted to adaptively control the importance of $\bm{O}'$ for instruction generation at the early stage. For the linear layers, we introduce the learnable scale vectors and use the dot product between the scale factors and the weight/bias tensor of each layer, respectively. In this way, we simplify the training process by retaining the parameters of LLaMA to stress the scene-instruction alignment and eliminate the potential risk of impairing the capacity of text generation~\cite{Lian_2022_SSF,jia2022visual}. The number of added parameters (\ie, $\rm \Psi$) only accounts for \textbf{7.2}\% of the entire model, demonstrating that \textsc{BEVInstructor} is a parameter-efficient framework.

\subsection{Instance-Guided Iterative Refinement}\label{sec:refinement}
Given the complexity of 3D environments, it is difficult to generate precise instructions that align with the scene layout. Humans usually describe a route by conceiving a rough draft based on the landmarks and then improve it~\cite{fernandes2023bridging,pan2023automatically}. Motivated by how humans refine their descriptions, we devise an instance-guided refinement strategy to learn from the generated landmarks and optimize the instructions. In the initial stage, \textsc{BEVInstructor} outputs a series of candidate instance words as initial landmark tokens $\mathcal{X}^I=\{\bm{x}_1^I,\bm{x}_2^I,\cdots\in \mathbb{R}^D\}$, \ie, $\mathcal{O}\times\mathcal{A}\rightarrow\mathcal{X}^I$. Next, the instance-guided draft is incorporated into the model to refine the instructions, \ie, $\mathcal{O}\!\times\!\mathcal{A}\!\times\!\mathcal{X}^I\!\rightarrow\!\mathcal{X}$. The optimization objective (Eq.~\ref{eq:autogre}) is reformulated as:
\begin{equation}
 \begin{aligned}
\max_{\rm \Theta}\big(\sum\nolimits_{l=1}^{L} \log P_{\rm \Theta}(\bm{x}_l|\bm{x}_{<l}, \mathcal{O}, \mathcal{A},\mathcal{I})+\sum\nolimits_{l=1}^{|\mathcal{X}^I|} \log P_{\rm \Theta}(\bm{x}_l^I|\bm{x}_{<l}^I, \mathcal{O}, \mathcal{A})\big).
\label{eq:refinement}
 \end{aligned}
\end{equation}
Parameter-efficient finetuning is omitted here for the sake of clarity. We implement multi-turn refinement in the process of generation (see Table~\ref{table:refinementablation}). Compared with existing methods~\cite{tan2019learning,wang2022counterfactual}, \textsc{BEVInstructor} can identify crucial objects based on informative perspective-BEV prompts. In this way, \textsc{BEVInstructor} enriches the object semantics in the generated instructions.

\subsection{Implementation Details}
\label{sec:implement}
\textsc{BEVInstructor} encodes the perspective embeddings and BEV embeddings by the visual encoder (\S\ref{sec:visualencoder}). Then the fused embeddings are served as the visual prompts for \textit{perspective-BEV prompt tuning} (\S\ref{sec:alignment}). Moreover, \textsc{BEVInstructor} employs a two-stage strategy for progressive instruction generation (\S\ref{sec:refinement})

\noindent\textbf{Visual Embedding.} The spatial shape of multi-view features is $H=W=14$. The orientation representation $\bm{\delta}_{t,k}$ of view $\bm{F}_{t,k}$ is defined as $(\text{cos}\varphi_{t,k}, \text{sin}\varphi_{t,k}$, $\text{cos}\phi_{t,k}, \text{sin}\phi_{t,k})$, where $\varphi$ and $\phi$ are the angles of heading and elevation, respectively. The embedding dimension is set as $D=768$. For BEV embeddings, the shape of the BEV plane (${H_b}\times{W_b}$) is set as $15\times 15$. The corresponding perception range is [-5.0 m, 5.0 m]. $N_\text{ref}=4$ reference points uniformly distributed over the height [-1.2 m, 2.0 m] are used for each BEV query. The relative 2D coordinates are used for position encoding $\bm{E}_p$. For BEV encoding, there are six deformable attention blocks in $\mathcal{F}^d$. The depth prediction network $\mathcal{E}^c$ employs discrete convolutions~\cite{philion2020lift,li2022bevdepth} (More detailed in Appendix).

\noindent\textbf{Word Embedding.} We follow the default tokenizer of LLaMA~\cite{touvron2023llama}. The word embedding dimension is set as $4096$. The max sequence token length is $128$.

\noindent\textbf{Network Architecture.} 
The transformer $\mathcal{F}^o$ has six blocks for perspective-BEV fusion. The lightweight transformer $\mathcal{Q}$ is used to map the BEV embedding into $N_q=10$ tokens, consisting of eight blocks. The backbone of LLM $\mathcal{F}^{\text{LM}}$ is initialized from LLaMA-7B~\cite{touvron2023llama} with $32$ layers. All experiments follow default training configurations on R2R~\cite{anderson2018vision}, REVERIE~\cite{reverie}, and UrbanWalk~\cite{huang2022assister}. Perspective-BEV prompts $\bm{O}'$ are inserted into the topmost $N_a=31$ layers.

\noindent\textbf{Finetuning.} We use the AdamW~\cite{loshchilov2017decoupled} as the optimizer with the learning rate of $1e^{-4}$ and overall batch size of eight with 20k iterations. We only finetune the small-scale parameters (\ie, $\rm \Theta^*\!\cup\!\Psi$ in Eq.~\ref{eq:tuning}, $<\!\!500$M) while freezing most parameters ($6.68$B) of \textsc{BEVInstructor}.

\noindent\textbf{Inference.} During inference, \textsc{BEVInstructor} processes a sequence of multi-view images to generate navigation instructions. This autoregressive method involves an iterative instruction generation process that builds upon previously generated instructions. Each new instruction is refined progressively using instance tokens until the <end> token.

\section{Experiments}
\label{sec:experiments}
\subsection{Experimental Settings}
\noindent\textbf{Datasets.} We conduct experiments on three datasets with instructions: R2R~\cite{anderson2018vision} and REVERIE~\cite{reverie} for indoor scenes, and UrbanWalk~\cite{huang2022assister} for outdoor scenes. 
\begin{itemize}[leftmargin=*]
	\setlength{\itemsep}{0pt}
	\setlength{\parsep}{-2pt}
	\setlength{\parskip}{-0pt}
	\setlength{\leftmargin}{-10pt}
	\item \noindent\textbf{R2R}~\cite{anderson2018vision} builds upon Matterport3D~\cite{Matterport3D}, including diverse photo-realistic house scenes. There are three splits for the experiment, \ie, \texttt{train} ($61$ scenes, $14,039$ instructions), \texttt{val seen} ($61$ scenes, $1,021$ instructions), and \texttt{val unseen} ($11$ scenes, $2,349$ instructions). There are three human-annotated navigation instructions for each path and the average length is approximately $29$ words. There are no overlapping scenes between \texttt{train} and \texttt{unseen} splits. Following previous studies~\cite{wang2022counterfactual,wang2023lana}, $6,482$K synthesized navigation route-instruction pairs from PREVALENT~\cite{hao2020towards} are also involved for training.

	\item \noindent\textbf{REVERIE}~\cite{reverie} extends the Matterport3D~\cite{Matterport3D} simulator to incorporate object annotations. It comprises indoor scenes with $4,140$ target objects and $21,702$ instructions with an average length of $18$ words. There are three splits for our experiment, \ie, \texttt{train} ($61$ scenes, $10,466$ instructions), \texttt{val seen} ($61$ scenes, $1,371$ instructions), and \texttt{val unseen} ($10$ scenes, $3,753$ instructions).

	\item \noindent\textbf{UrbanWalk}~\cite{huang2022assister} involves outdoor scenes from the simulator with $26,808$ naturalistic instructions. On average, there are 21.7 words per instruction.
\end{itemize}
\noindent\textbf{Evaluation Metrics.} Following previous studies~\cite{agarwal2019visual,wang2022counterfactual,wang2023lana}, we employ five standard metrics: \textbf{1) BLEU}~\cite{papineni2002bleu} refers to the geometric mean of $n$-gram precision scores computed over reference and candidate descriptions. \textbf{2) CIDEr}~\cite{vedantam2015cider} represents the average cosine similarity between $n$-grams of the reference and candidate descriptions, weighted by their corresponding term frequency-inverse document frequency values. \textbf{3) METEOR}~\cite{banerjee2005meteor} is defined as the harmonic mean of precision and recall of unigram matches between sentences. \textbf{4) ROUGE}~\cite{lin2004rouge} is a measure of correspondence between the reference and candidate texts by computing the recall and precision scores for each $n$-gram size, word sequences and word pairs, and thus averaging them by a weighted F-measure. \textbf{5) SPICE}~\cite{anderson2016spice} is the F-score of the scenegraph~\cite{schuster2015generating} tuples of the candidate sentence and all reference sentences. \textbf{SPICE} is considered as the primary indicator.

\noindent\textbf{Reproducibility.} \textsc{BEVInstructor} is implemented in PyTorch and all models are trained and tested using a single machine with 2 NVIDIA A40 GPUs.

\begin{table*}[!bth]
	\centering
	\captionsetup{font=small}
	\caption{\small Quantitative comparison results for \textbf{instruction generation} on R2R~\cite{anderson2018vision} \texttt{val\!\! seen} and \texttt{val\!\! unseen}. See~\S\ref{sec:qresults1} for more details.}
	\label{table:R2Rig}
		\resizebox{1\textwidth}{!}{
		\setlength\tabcolsep{1pt}
		\renewcommand\arraystretch{1.0}
	\begin{tabular}{|rl||cccccc|cccccc|}
	\hline \thickhline
	\rowcolor{mygray}
	~ & & \multicolumn{6}{c|}{R2R \texttt{val} \texttt{seen}} & \multicolumn{6}{c|}{R2R \texttt{val} \texttt{unseen}}\\
	\cline{3-14}\cline{3-14}
	\multicolumn{2}{|c||}{\multirow{-2}{*}{\cellcolor{gray!25}Methods}}
	&\textbf{\texttt{SPICE}}\!~$\uparrow$ &\texttt{Bleu-1}\!~$\uparrow$ &\texttt{Bleu-4}\!~$\uparrow$ &\texttt{CIDEr}\!~$\uparrow$ &\texttt{Meteor}\!~$\uparrow$ &\texttt{Rouge}\!~$\uparrow$  &\textbf{\texttt{SPICE}}\!~$\uparrow$ &\texttt{Bleu-1}\!~$\uparrow$ &\texttt{Bleu-4}\!~$\uparrow$ &\texttt{CIDEr}\!~$\uparrow$ &\texttt{Meteor}\!~$\uparrow$ &\texttt{Rouge}\!~$\uparrow$  \\
	\hline
	\hline
	BT-speaker~\cite{fried2018speaker}&\!\!\pub{NeurIPS2018}    &0.182 &0.685 &0.253 &0.483 &0.227 &0.473   &0.178 &0.658 &0.250 &0.391 &0.209 &0.440 \\
    EDrop-speaker~\cite{tan2019learning}&\!\!\pub{NAACL2019}   &0.195 &0.701 & 0.265 &0.486 &0.224 &0.463   &0.184 & 0.660 &0.260 &0.413 &0.215 &0.455 \\
	CCC-speaker~\cite{wang2022counterfactual}&\!\!\pub{CVPR2022}  &{0.196}  &{0.698} & 0.267 & 0.498 & 0.233 & 0.467  &0.183 &0.679 & 0.254 & 0.401 &0.226 &0.456 \\
	Lana~\cite{wang2023lana}&\!\!\pub{CVPR2023}   &0.201 &0.694 &0.270 &0.503 &0.230 &0.473   &0.194 &0.689 &0.260 &0.419 &0.219 &0.463  \\
  \hline
  \textsc{BEVInstructor} &\textbf{(Ours)} & \textbf{0.220}  & \textbf{0.731} & \textbf{0.285} & \textbf{0.549} & \textbf{0.238} & \textbf{0.480} & \textbf{0.208} & \textbf{0.699} & \textbf{0.264} & \textbf{0.449} & \textbf{0.230} & \textbf{0.467}  \\
  \hline
	\end{tabular}
  }
\end{table*}
\begin{table*}[!bth]
	\centering
	\captionsetup{font=small}
	\caption{\small Quantitative comparison results for \textbf{instruction generation} on REVERIE~\cite{reverie} \texttt{val\!\! seen} and \texttt{val\!\! unseen}. See~\S\ref{sec:qresults1} for more details.}
	\label{table:REVERIEig}
		\resizebox{1\textwidth}{!}{
		\setlength\tabcolsep{1pt}
		\renewcommand\arraystretch{1.0}
	\begin{tabular}{|rl||cccccc|cccccc|}
	\hline \thickhline
	\rowcolor{mygray}
	~ & & \multicolumn{6}{c|}{REVERIE \texttt{val} \texttt{seen}} & \multicolumn{6}{c|}{REVERIE \texttt{val} \texttt{unseen}}\\
	\cline{3-14}\cline{3-14}
	\multicolumn{2}{|c||}{\multirow{-2}{*}{\cellcolor{gray!25}Methods}}
	&\textbf{\texttt{SPICE}}\!~$\uparrow$ &\texttt{Bleu-1}\!~$\uparrow$ &\texttt{Bleu-4}\!~$\uparrow$ &\texttt{CIDEr}\!~$\uparrow$ &\texttt{Meteor}\!~$\uparrow$ &\texttt{Rouge}\!~$\uparrow$  &\textbf{\texttt{SPICE}}\!~$\uparrow$ &\texttt{Bleu-1}\!~$\uparrow$ &\texttt{Bleu-4}\!~$\uparrow$ &\texttt{CIDEr}\!~$\uparrow$ &\texttt{Meteor}\!~$\uparrow$ &\texttt{Rouge}\!~$\uparrow$  \\
	\hline
	\hline
	BT-speaker~\cite{fried2018speaker}&\!\!\pub{NeurIPS2018}    &0.121 &0.693 &0.347 &0.269 &0.223 &0.602  &0.103 &0.664 &0.302 &0.190 &0.200 &0.519 \\
    EDrop-speaker~\cite{tan2019learning}&\!\!\pub{NAACL2019}   &0.133 &0.666 & 0.353 &0.517 &0.237 &0.589   &0.114 & 0.655 &0.312 &0.252 &0.222 &0.534 \\
	CCC-speaker~\cite{wang2022counterfactual}&\!\!\pub{CVPR2022}  &0.138  &0.680 & 0.377 & 0.549 &0.244 &0.593  &0.117 &0.671 & 0.329 & 0.280 &0.233 &0.533 \\
	Lana~\cite{wang2023lana}&\!\!\pub{CVPR2023}   &0.137 &0.714 &0.408 &0.619 &0.280 &0.615   &0.108 &0.701 &0.332 &0.406 &0.237 &0.542  \\
  \hline
	\textsc{BEVInstructor} &\textbf{(Ours)} & \textbf{0.208}  & \textbf{0.773} & \textbf{0.425} & \textbf{0.745} & \textbf{0.324} & \textbf{0.635} & \textbf{0.159} & \textbf{0.732} & \textbf{0.335} & \textbf{0.489} & \textbf{0.267} & \textbf{0.560}  \\
  \hline
	\end{tabular}
  }
\end{table*}

\subsection{Quantitative Results}
\label{sec:qresults1}

\noindent\textbf{R2R.} As depicted in Table~\ref{table:R2Rig}, \textsc{BEVInstructor} achieves the best performance across all metrics on both \texttt{val} splits. Notably, on the primary metrics --- \textbf{SPICE}, it achieves an improvement of \textbf{1.9\%} on \texttt{val\!\! seen} and \textbf{1.4\%} on \texttt{val\!\! unseen}. Additionally, our model surpasses the leading benchmarks of prior studies, achieving a \textbf{4.6\%} improvement on \texttt{CIDEr} of R2R \texttt{val seen} split and a \textbf{3.0\%} enhancement on \texttt{CIDEr} of R2R \texttt{val unseen} split. This verifies the efficacy of \textsc{BEVInstructor} in generating navigation instructions for indoor scenarios.

\noindent\textbf{REVERIE.} Table~\ref{table:REVERIEig} compares \textsc{BEVInstructor} with the recent state-of-the-art instruction generation models~\cite{fried2018speaker,tan2019learning,wang2022counterfactual,wang2023lana} on REVERIE dataset. \textsc{BEVInstructor} outperforms previous approaches across all the evaluation metrics on the \texttt{val} split. Specifically, on the \texttt{val seen} split, \textsc{BEVInstructor} exceeds the previous best model by \textbf{7.0\%} on \texttt{SPICE}, \textbf{12.6\%} on \texttt{CIDEr}, and \textbf{4.4\%} on \texttt{Meteor}. On the \texttt{val unseen} split, \textsc{BEVInstructor} improves the performance by \textbf{4.2\%} on \texttt{SPICE}, \textbf{8.3\%} on \texttt{CIDEr}, and \textbf{3.0\%} on \texttt{Meteor}. These results underscore the versatility and generality of our architectural design for goal-based tasks.

\setlength{\intextsep}{1pt}
\begin{wraptable}{r}{6.5cm}
	\centering
	\captionsetup{font=small}
	\caption{\small{Quantitative comparison results for \textbf{instruction generation} on UrbanWalk~\cite{huang2022assister} \texttt{test}. See~\S\ref{sec:qresults1} for more details.}}
	\label{table:urbanig}
		\resizebox{0.53\textwidth}{!}{
		\setlength\tabcolsep{0.2pt}
		\renewcommand\arraystretch{1}
	\begin{tabular}{|rl||c c c c|}
	\hline \thickhline
	\rowcolor{mygray}
	~ & & \multicolumn{4}{c|}{UrbanWalk \texttt{test}} \\
	\cline{3-6}
	\multicolumn{2}{|c||}{\multirow{-2}{*}{\cellcolor{gray!25}Methods}}
	&\textbf{\texttt{SPICE}}\!~$\uparrow$  &\texttt{Bleu-4}\!~$\uparrow$ &\texttt{Meteor}\!~$\uparrow$ &\texttt{Rouge}\!~$\uparrow$\\
	\hline
	\hline
	BT-speaker~\cite{fried2018speaker}&\!\!\pub{NeurIPS2018}   &0.524 & 0.408 &0.350 &0.620 \\
    EDrop-speaker~\cite{tan2019learning}&\!\!\pub{NAACL2019}    &0.531 &0.435 &0.358 &0.634  \\
	ASSISTER~\cite{huang2022assister}&\!\!\pub{ECCV2022}  &0.451  &0.164 & 0.319 &0.557  \\
	Kefa-speaker~\cite{zeng2023kefa}&\!\!\pub{Arxiv2023} &0.566  &0.450 &0.378 &0.655 \\
  \hline
  \textsc{BEVInstructor} &\textbf{(Ours)} & \textbf{0.679} & \textbf{0.575} & \textbf{0.451} & \textbf{0.786}   \\
  \hline
	\end{tabular}
  }
\end{wraptable}

\noindent\textbf{UrbanWalk.} Table~\ref{table:urbanig} presents comparison results on UrbanWalk dataset. On UrbanWalk test split, \textsc{BEVInstructor} outperforms previous methods by a significant advancement, \textbf{11.3\%} on \texttt{SPICE}, \textbf{12.5\%} on \texttt{Bleu-4}, \textbf{7.3\%} on \texttt{Meteor} and \textbf{13.1\%} on \texttt{Rouge}. This suggests \textsc{BEVInstructor} is also capable of handling more challenging outdoor scenes.

\subsection{Qualitative Results}
\label{sec:resultfig}

\begin{figure*}[t]
	  \centering
		  \includegraphics[width=0.95\linewidth]{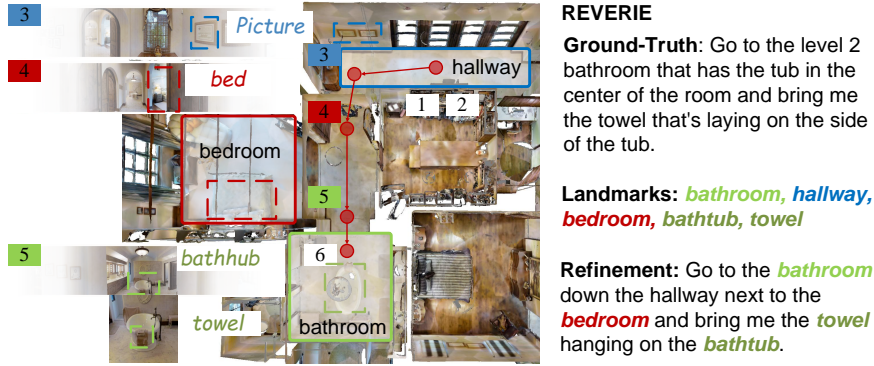}
	\captionsetup{font=small}
	\caption{\small Visual comparison results between ground truth and \textsc{BEVInstructor} for instruction generation on REVERIE~\cite{reverie}. See \S\ref{sec:resultfig} for more details.}
	\label{fig:qualitative}
\end{figure*}

Fig.\!~\ref{fig:qualitative} provides qualitative comparisons of \textsc{BEVInstructor} against the ground truth on the REVERIE. \textsc{BEVInstructor} shows an enhanced capability in identifying scenes and objects related to action views, and explicitly incorporates these elements into the instructions in the refinement stage.

\subsection{Diagnostic Experiment}\label{sec:diagnostic}
To assess the efficacy of essential modules of \textsc{BEVInstructor}, we conduct a series of detailed ablation studies on \texttt{val unseen} split of R2R~\cite{anderson2018vision}.

\begin{table*}[t]
	\centering
	\captionsetup{font=small}
		\caption{\small Ablation study on R2R~\cite{anderson2018vision} \texttt{val\!\! unseen}. See \S\ref{sec:diagnostic} for more details.}
		\label{table:diagnostic}
		\resizebox{0.9\textwidth}{!}{
		\setlength\tabcolsep{3pt}
		\renewcommand\arraystretch{1.0}
	\begin{tabular}{|l|c|c|c|c|cccccc|}
	\hline \thickhline
	\rowcolor{mygray}
	~ & & & & & \multicolumn{6}{c|}{R2R \texttt{val} \texttt{unseen}}\\
	\cline{6-11}\cline{6-11}
	\multirow{-2}{*}{\cellcolor{gray!25}\#} & \multirow{-2}{*}{\cellcolor{gray!25}Perspective} & \multirow{-2}{*}{\cellcolor{gray!25}BEV} & \multirow{-2}{*}{\cellcolor{gray!25}Fusion} & \multirow{-2}{*}{\cellcolor{gray!25}Refinement} &\textbf{\texttt{SPICE}}\!~$\uparrow$ &\texttt{Bleu-1}\!~$\uparrow$ &\texttt{Bleu-4}\!~$\uparrow$ &\texttt{CIDEr}\!~$\uparrow$ &\texttt{Meteor}\!~$\uparrow$ &\texttt{Rouge}\!~$\uparrow$  \\
	\hline
	\hline
	1 & \cmark & 		&       & &0.154 &0.625 &0.170 &0.209 &0.198 &0.392 \\
	2 & 	   & \cmark &       & &0.172  &0.653  &0.184  &0.281  &0.206  &0.405  \\
    3 & \cmark & \cmark &     	& &0.180 &0.673 &0.217 &0.342 &0.224 &0.442 \\
	4 & \cmark & \cmark & \cmark& &0.190 &0.683 &0.238 &0.373 &0.224 &0.453  \\
	5 & \cmark & \cmark &     	& \cmark &0.192 & 0.676 &0.242 &0.419 &0.220 &0.455 \\
	6 & \cmark & \cmark & \cmark& \cmark &\textbf{0.208} & \textbf{0.699} & \textbf{0.264} & \textbf{0.449} & \textbf{0.230} & \textbf{0.467} \\
  \hline
	\end{tabular}
  }
\end{table*}

\noindent\textbf{Overall Design.} We first study the efficacy of the core components of \textsc{BEVInstructor} in Table~\ref{table:diagnostic}. Row \#1 illustrates the impact of fine-tuning MLLMs. This shows competitive performance, demonstrating its potential by elevating language capabilities. Row \#2 and \#3 indicate that the integration of BEV features alongside perspective features yields notable performance improvements by \textbf{6.1\%} on \texttt{CIDEr}. 
From row \#3 and \#4, compared with simply concatenating features, fusing BEV and perspective features through the transformer module results in a greater performance improvement by \textbf{1.0\%} on \texttt{SPICE}. Comparisons between row \#3 and \#5, as well as row \#4 and \#6, underscore the efficacy of the instance-guided iterative refinement module. In row \#6, we combine all the components, and obtain the best performance. This$_{\!}$ suggests$_{\!}$ that$_{\!}$ these$_{\!}$ modules$_{\!}$ are$_{\!}$ complementary$_{\!}$ to$_{\!}$ each$_{\!}$ other,$_{\!}$ and$_{\!}$ confirms$_{\!}$ the$_{\!}$ effectiveness$_{\!}$ of$_{\!}$ our$_{\!}$ whole$_{\!}$ design.

\begin{table}[t]
	\captionsetup{font=small}
	\caption{\small{Ablation study of other transformer-based algorithms on R2R val unseen. See \S\ref{sec:diagnostic} for more details.}}
	\label{table:R2RLLMig}
	\centering
		\resizebox{0.8\textwidth}{!}{
		\setlength\tabcolsep{2pt}
		\renewcommand\arraystretch{1.0}
  \hspace{-16pt}
	\begin{tabular}{|rl||cccccc|}
	\hline \thickhline
	\rowcolor{mygray}
	~ & & \multicolumn{6}{c|}{R2R \texttt{val} \texttt{unseen}}\\
	\cline{3-8}\cline{3-8}
	\multicolumn{2}{|c||}{\multirow{-2}{*}{\cellcolor{gray!25}Methods}}
	&\textbf{\texttt{SPICE}}\!~$\uparrow$ &\texttt{Bleu-1}\!~$\uparrow$ &\texttt{Bleu-4}\!~$\uparrow$ &\texttt{CIDEr}\!~$\uparrow$ &\texttt{Meteor}\!~$\uparrow$ &\texttt{Rouge}\!~$\uparrow$  \\
	\hline
	\hline
	GPT-4V$_{\!}$~\cite{openai2023gpt4}&\!\!\pub{Arxiv2023} &0.098 &0.403 &0.079 &0.076 &0.130 &0.296 \\
	AutoVLN(GPT2)$_{\!}$~\cite{chen2022learning}&\!\!\pub{ECCV2022} &0.145 &0.613 &0.181 &0.248 &0.188 &0.398 \\
	PASTS$_{\!}$~\cite{wang2024pasts}&\!\!\pub{EAAI2024} &0.151  &0.645  &0.195  &0.258  &0.186  &0.415  \\
  	InstructBLIP$_{\!}$~\cite{dai2023instructblip}&\!\!\pub{NeurIPS2023} &0.163 &0.666 &0.220  &0.321  &0.201 &0.418   \\
  \hline
  \textsc{BEVInstructor}\!\! &\textbf{(Ours)} & \textbf{0.208} & \textbf{0.699} & \textbf{0.264} & \textbf{0.449} & \textbf{0.230} & \textbf{0.467}  \\
  \hline
	\end{tabular}
  }
\end{table}
\begin{table}[t]
	\caption{A set of ablation studies on R2R$_{\!}$~\cite{anderson2018vision} \texttt{val unseen}. See \S\ref{sec:diagnostic} for more details.}
	\hspace{-0.7em}
	\begin{subtable}{.475\linewidth}
		\captionsetup{width=.95\linewidth}
		\resizebox{\textwidth}{!}{
			\setlength\tabcolsep{1pt}
			\renewcommand\arraystretch{1.35}
			\begin{tabular}{c||c c c c c}
				\thickhline
				{\cellcolor{gray!25}Fusion} & \textbf{\texttt{SPICE}}\!~$\uparrow$ &\texttt{Bleu-4}\!~$\uparrow$ &\texttt{CIDEr}\!~$\uparrow$ &\texttt{Meteor}\!~$\uparrow$ &\texttt{Rouge}\!~$\uparrow$ \\
				\hline\hline
				Addition &0.185  &0.226  &0.366  &0.214  &0.450 \\
				Concat  &0.184   &0.192  &0.310  &0.213 &0.436 \\
				\hline
				\textbf{Ours}  & \textbf{0.208} & \textbf{0.264} & \textbf{0.449} & \textbf{0.230} & \textbf{0.467} \\ 				\hline
			\end{tabular}
		}
		\setlength{\abovecaptionskip}{0.35cm}
		\setlength{\belowcaptionskip}{-0.2cm}
		\caption{\small{Ablation study of different fusion of Perspective-BEV features on R2R~\cite{anderson2018vision} \texttt{val unseen}.}}
		\label{table:fusionablation}
	\end{subtable}
	\begin{subtable}{.475\linewidth}
		\captionsetup{width=.95\linewidth}
		\resizebox{\textwidth}{!}{
			\setlength\tabcolsep{1pt}
			\renewcommand\arraystretch{1.2}
			\begin{tabular}{c||c c c c c}
				\thickhline
				 {\cellcolor{gray!25} steps}
     &
     \textbf{\texttt{SPICE}}\!~$\uparrow$  &\texttt{Bleu-4}\!~$\uparrow$ &\texttt{CIDEr}\!~$\uparrow$ &\texttt{Meteor}\!~$\uparrow$ &\texttt{Rouge}\!~$\uparrow$  \\
				\hline
				\hline
				Base &0.190 &0.238 &0.373 &0.224 &0.453 \\
				One   & \textbf{0.208}  & \textbf{0.264} & 0.449 & \textbf{0.230} & 0.467 \\
				Two	 &0.204 &\textbf{0.264}  &\textbf{0.456}  &\textbf{0.230}  &\textbf{0.474}  \\
				\hline
			\end{tabular}
		}
		\setlength{\abovecaptionskip}{0.4cm}
		\setlength{\belowcaptionskip}{-0.2cm}
		\caption{\small{Ablation study of different steps of refinement on R2R~\cite{anderson2018vision} \texttt{val unseen}. \\} 
}
		\label{table:refinementablation}
	\end{subtable}
\end{table}

\noindent\textbf{Language Architectures.} Table~\ref{table:R2RLLMig} presents the  performance comparison of various transformer-based algorithms on R2R val unseen split. Except for GPT-4V, the other methods are fine-tuned on R2R train split. The results show that more advanced language architectures can effectively adapt to the task of generating navigation instructions with just fine-tuning, achieving competitive performance.$_{\!}$ This$_{\!}$ confirms$_{\!}$ the$_{\!}$ potential$_{\!}$ of$_{\!}$ MLLMs$_{\!}$ to$_{\!}$ enhance$_{\!}$ instruction$_{\!}$ generation.

\noindent\textbf{Fusion of Perspective-BEV Features.} We compare the performance of three different approaches for fusing perspective and BEV features: \textbf{i)} addition of perspective and BEV features, \textbf{ii)} concatenation of perspective features and BEV features along the token dimension, and \textbf{iii)} fusion by transformer modules. The results are summarized in Table~\ref{table:fusionablation}. Note that, the fusion design of \textsc{BEVInstructor} outperforms the other two simpler fusion approaches. This robustly validates the architecture design of \textsc{BEVInstructor}.

\noindent\textbf{Refinement.} Table~\ref{table:refinementablation} presents the performance comparison of different refinement steps on the R2R \texttt{val unseen} split. As shown in row \#1 and \#2, the instance-guided iterative refinement improves the instructions through multi-step reasoning. However, from row \#2 and \#3, further increasing the steps of refinements only brings limited improvement.

\begin{table}[!bth]
	\caption{A set of analysis of instruction quality. See \S\ref{sec:quality} for more details.}
	\hspace{-0.7em}
	\centering
	  \begin{subtable}{0.475\linewidth}
		\centering
		\resizebox{1\textwidth}{!}{
			\setlength\tabcolsep{2pt}
			\renewcommand\arraystretch{1.18}
		\begin{tabular}[!t]{rl||c c|c c}
			\hline \thickhline

			\rowcolor{mygray}
			~ & & \multicolumn{2}{c|}{HAMT~\cite{chen2021history}} &  \multicolumn{2}{c}{DUET~\cite{chen2022think}}\\
			\cline{3-6}
			\multicolumn{2}{c||}{\multirow{-2}{*}{\cellcolor{mygray} Method}}
			&\texttt{SR}\!~$\uparrow$  & \multicolumn{1}{c|}{\texttt{SPL}\!~$\uparrow$} &\texttt{SR}\!~$\uparrow$ &\texttt{SPL}\!~$\uparrow$\\
			\hline
			\hline
			\rowcolor{mygray}
			Human annotation~\cite{reverie}&\!\!\pub{CVPR2020} &32.95 &30.20 &46.98 &33.73 \\
			BT-speaker~\cite{fried2018speaker}&\!\!\pub{NeurIPS2018} &22.48 &19.47 &28.41 &15.30 \\
			EDrop-speaker~\cite{tan2019learning}&\!\!\pub{NAACL2019} &23.74 &20.98 &30.66 &19.27 \\
			CCC-speaker~\cite{wang2022counterfactual}&\!\!\pub{CVPR2022} &23.80 &21.18 &28.84 &14.36 \\
			Lana~\cite{wang2023lana}&\!\!\pub{CVPR2023} &23.94 &21.34 &31.61 &21.26 \\
			\hline
			\textsc{BEVInstructor}\!\! &\textbf{(Ours)} &\textbf{25.68} &\textbf{22.48} &\textbf{33.81} &\textbf{23.23}  \\
			\hline
		\end{tabular}
		}
		\captionsetup{font=small}
		\caption{\small{The performance of HAMT~\cite{chen2021history} and DUET~\cite{chen2022think} guided by instructions generated on REVERIE~\cite{reverie} \texttt{val unseen}.}}
		\label{table:pathguide}
	\end{subtable}
	\begin{subtable}{0.475\linewidth}
		\centering
		\resizebox{1\textwidth}{!}{
			\setlength\tabcolsep{2pt}
			\renewcommand\arraystretch{1.03}
		\begin{tabular}[!t]{rl||c c c c}
			\hline \thickhline
			\rowcolor{mygray}
			~ & & \multicolumn{4}{c}{R2R \texttt{val unseen}} \\
			\cline{3-6}
			\multicolumn{2}{c||}{\multirow{-2}{*}{\cellcolor{gray!25}Data Source}}
			&\texttt{TL}  &\texttt{NE}\!~$\downarrow$ &\texttt{SR}\!~$\uparrow$ &\texttt{SPL}\!~$\uparrow$\\
			\hline
			\hline
			Original~\cite{reverie}&\!\!\pub{CVPR2020} &9.62 &5.86 &45.4 &41.8 \\
			+ BT-speaker~\cite{fried2018speaker}&\!\!\pub{NeurIPS2018} &9.81 &5.95 &45.1 &41.5 \\
			+ EDrop-speaker~\cite{tan2019learning}&\!\!\pub{NAACL2019} &9.33 &5.68 &45.5 &42.2 \\
			+ CCC-speaker~\cite{wang2022counterfactual}&\!\!\pub{CVPR2022} &9.43 &5.73 &45.3 &42.0 \\
			+ Lana~\cite{wang2023lana}&\!\!\pub{CVPR2023} &9.48 &5.75 &45.6 &42.1 \\
			\hline
			\textsc{BEVInstructor}\!\! &\textbf{(Ours)} &9.96 &\textbf{5.66} &\textbf{47.1} &\textbf{43.6}  \\
			\hline
		\end{tabular}
		}
		\captionsetup{font=small}
	\caption{\small{The$_{\!}$ results$_{\!}$ of$_{\!}$ EDrop-follower~\cite{tan2019learning}$_{\!}$ using$_{\!}$ different$_{\!}$ generators$_{\!}$ for$_{\!}$ data$_{\!}$ augmentation$_{\!}$ on$_{\!}$ R2R~\cite{anderson2018vision}$_{\!}$ \texttt{val$_{\!}$ unseen}.}}
	\label{table:dataaug}
	\end{subtable}

\end{table}

\subsection{Instruction Quality Analysis}
\label{sec:quality}
The above captioning metrics reflect the word alignment quality of the generated instructions. 
To further demonstrate the effectiveness of the instructions, we conduct the following experiments to evaluate the alignment of instructions and trajectories in actual vision-language navigation tasks. We compare two aspects, \ie, Path Guiding Proficiency, and Data Augmentation, with previous methods to further validate the performance of \textsc{BEVInstructor}.

\noindent\textbf{Path Guiding Proficiency.} Table~\ref{table:pathguide} presents the comparison performance of HAMT~\cite{chen2021history} and DUET~\cite{chen2022think} guided by instructions from different generators on REVERIE \texttt{val unseen}. Concretely, we reproduce \textbf{one} instruction for each path on REVERIE \texttt{val unseen} and replace the original instructions with them. 
We follow~\cite{chen2021history,chen2022think,wang2023lana,wang2022counterfactual} and adopt the Success Rate (SR) and Success rate weighted by Path Length (SPL) of navigation as metrics. \textsc{BEVInstructor} outperforms others across all metrics of HAMT and DUET. These$_{\!}$ results$_{\!}$ confirm$_{\!}$ the$_{\!}$ fidelity$_{\!}$ and$_{\!}$ generalization$_{\!}$ of$_{\!}$ our$_{\!}$ instructions$_{\!}$.

\noindent\textbf{Data Augmentation.} One application of the navigation instruction generator is to create diverse instructions for data augmentation in Vision-Language Navigation (VLN) agents. We randomly sample paths in the R2R \texttt{train} split and employ different generators to produce instructions. The generated instructions combined with the original ones are used to train the EDrop-follower~\cite{tan2019learning}. In Table~\ref{table:dataaug}, the instructions generated by \textsc{BEVInstructor} for data augmentation enhance the performance of VLN, leading to promising improvements.  

\section{Conclusion}
\label{sec:conclusion}
Navigation instruction generation has been of great importance in many disciplines. However, existing studies encounter the following challenges: \textbf{i)} they rely exclusively on perspective features, ignoring the geometric prior and object semantics inherent in 3D scenes, and \textbf{ii)} their language decoders are limited by a lack of extensive prior world knowledge and the scale of the model. In light of this, we propose \textsc{BEVInstructor} that integrates BEV features with MLLMs to jointly improve 3D perception and linguistic capabilities. \textsc{BEVInstructor} exhibits superior performance in comparison to previous studies. This work brings us closer to developing interactive and trustworthy navigation robots.

%
%
\bibliographystyle{splncs04}
\bibliography{main}

\newpage
\appendix
\setcounter{section}{0}
\setcounter{table}{0}
\setcounter{figure}{0}
\renewcommand\thesection{\Alph{section}}
\renewcommand{\thetable}{S\arabic{table}}
\renewcommand{\thefigure}{S\arabic{figure}}

\makesupptitle{Supplementary Material}

This document provides more details of \textsc{BEVInstructor} as follows:
    \begin{itemize}
    \setlength{\itemsep}{0pt}
        \item  \textit{Implementation Details of \textsc{BEVInstructor}(\S\ref{sec:details}).} We introduce more implementation details of \textsc{BEVInstructor} and provide the pseudo-code for the training procedure.
        \item  \textit{Additional Quantitative and Qualitative Results (\S\ref{sec:supplequan}).} More quantitative and qualitative results are provided. 
        \item \textit{Discussion (\S\ref{sec:discussion}).} We offer further discussion on the limitations, social impact, and future work of \textsc{BEVInstructor}.
 \end{itemize}

\section{Implementation Details of \textsc{BEVInstructor}} 
\label{sec:details}
\noindent\textbf{BEV Encoding with 3D Detection.} To establish a holistic 3D scene understanding, 3D detection is used to supervise the BEV encoding~\cite{wang2023embodiedscan,li2022bevformer} in Eq.~\ref{eq:bevem}. For 3D detection, we adopt the 3D bounding boxes~\cite{liu2023bird} of Matterport3D~\cite{Matterport3D} following the same \textit{seen}/\textit{unseen} splits as R2R~\cite{anderson2018vision}. It contains 17 categories, covering common objects in daily life. The bipartite matching and the bounding box losses~\cite{li2022bevformer} are employed for detection. The BEV encoder is optimized by AdamW\cite{loshchilov2017decoupled} for 500 epochs with a learning rate of $1\!\times\!10^{-4}$.

In Table~\ref{table:detection}, we provide the detailed results of the main categories. It is noted that we adhere to the same settings as those used in  BEVFormer~\cite{li2022bevformer}, \ie, the shape of the BEV plane, the perception range, and the distribution of reference points (\S\ref{sec:implement}). For metrics, we use the 3D IoU-based mean average precision (mAP) with thresholds of 0.5. It demonstrates our depth consistency weight (Eq.~\ref{eq:bevweight}) facilitates the BEV projection and obtains higher quality BEV representations for scene perception.

We also present the pseudo-code of the training procedure in Algorithm~\ref{alg:bevinst}.

\begin{table*}[!bth]
	\centering
			\resizebox{0.98\textwidth}{!}{
			\setlength\tabcolsep{3pt}
			\renewcommand\arraystretch{1.0}
	\begin{tabular}{c||c|ccccccccccc}
	\hline \thickhline
	\rowcolor{mygray}
	Models &\small{mAP$\uparrow$}
	&\footnotesize{bed}
    &\footnotesize{table}
    &\footnotesize{door}
    &\footnotesize{sofa}
	&\footnotesize{chair}
    &\footnotesize{shelving}
	&\footnotesize{cabinet}
	&\footnotesize{plant}
	&\footnotesize{sink}
    &\footnotesize{cushion}
	&\footnotesize{monitor} \\
	\hline
	\hline
	BEVFormer~\cite{li2022bevformer} &23.37  &33.75 &32.71 &15.27 &29.55  &30.15 &7.59 &26.07 &21.96 &21.63 &17.23 &21.14  \\
	\hline
	\textbf{\textsc{BEVInstructor}(Ours)}  &\textbf{25.42}  &39.02 &37.20 &16.89 &31.09  &31.26 &9.44 &31.96 &24.65 &18.58 &21.53 &18.04 \\
	\hline
	\end{tabular}
	}
	\captionsetup{font=small}
		\caption{\small{Detailed results of the main categories on the \textit{unseen} scenes.}}
		\label{table:detection}
\end{table*}

\begin{algorithm}
    \caption{The pseudo-code of training for \textsc{\textsc{BEVInstructor}}.}\label{alg:bevinst}
	\textbf{Arguments:}
	Multi-view Image Features $\{\bm{F}_{t,k}\}_{k=1}^K$ with Orientation Angles $\{\delta_{t,k}\}_{k=1}^K$,
	Perspective Embedding $\bm{P}_t$,
	Action Embedding $\bm{a}_t$, BEV Embedding $\bm{B}_t$, Complete Observation Embedding $\bm{O}_t$,
	Instruction Tokens $\mathcal{X}\!=\!\{\bm{x}_l\}_{l=1}^{L}$, Landmark Tokens $\mathcal{X}^I$, the maximum iteration $N$, Perspective Encoder $\mathcal{E}^{\bm{P}\!}$, BEV Encoder $\mathcal{E}^{\bm{B}\!}$, Perspective-BEV Encoder $\mathcal{E}^{\bm{O}\!}$, MLLM $\mathcal{E}^{LLM\!}$.
    \begin{algorithmic}[1]
    \State Initialize $\mathcal{E}^{\bm{P}\!}$, $\mathcal{E}^{\bm{B}\!}$, $\mathcal{E}^{\bm{O}\!}$, $\mathcal{E}^{LLM\!}$
    \For{iteration $i \in [1,\dots,N]$}
		\State Sample a pretraining task $\mathcal{T}$ from \{\textit{Landmarks}, \textit{Instructions}\}
		\State $[\bm{P}_t, \bm{a}_t] = \mathcal{E}^{\bm{P}}(\bm{F}_{t,k}, \delta_{t,k})$ \Comment{Defined in Eq.~\ref{eq:panoemb}, \ref{eq:actemb}.}
		
		\State $\bm{B}_t = \mathcal{E}^{\bm{B}}(\bm{F}_{t,a})$ \Comment{Defined in Eq.~\ref{eq:bevem}, \ref{eq:bevweight}, \ref{eq:fusion}.}

		\State $\bm{O}_t = \mathcal{E}^{\bm{O}}(\bm{B}_{t}, [\bm{P}_t, \bm{a}_t])$ \Comment{Defined in Eq.~\ref{eq:bevprompt}, \ref{eq:tuning}.}

		\If{$\mathcal{T}$ is \textit{Landmarks}}
			\State $\mathcal{X}^I = \mathcal{E}^{LLM}(\bm{x}_{<l}, \bm{O}_{1:T})$ \Comment{Defined in Eq.~\ref{eq:llminput}.}
		\Else{$\mathcal{T}$ is \textit{Instructions}}
			\State $\mathcal{X} = \mathcal{E}^{LLM}(\bm{x}_{<l}, \bm{O}_{1:T}, \mathcal{X}^I)$ \Comment{Defined in Eq.~\ref{eq:refinement}.}
		\EndIf

		\State Update $\mathcal{E}^{\bm{P}\!}$, $\mathcal{E}^{\bm{B}\!}$, $\mathcal{E}^{\bm{O}\!}$, $\mathcal{E}^{LLM\!}$
		\EndFor
		\Return $\mathcal{E}^{\bm{P}\!}$, $\mathcal{E}^{\bm{B}\!}$, $\mathcal{E}^{\bm{O}\!}$, $\mathcal{E}^{LLM\!}$
    \end{algorithmic}
\end{algorithm}

\section{Additional Quantitative and Qualitative Results} \label{sec:supplequan}

\begin{figure*}[!bth]
	  \centering
		  \includegraphics[width=0.92\linewidth]{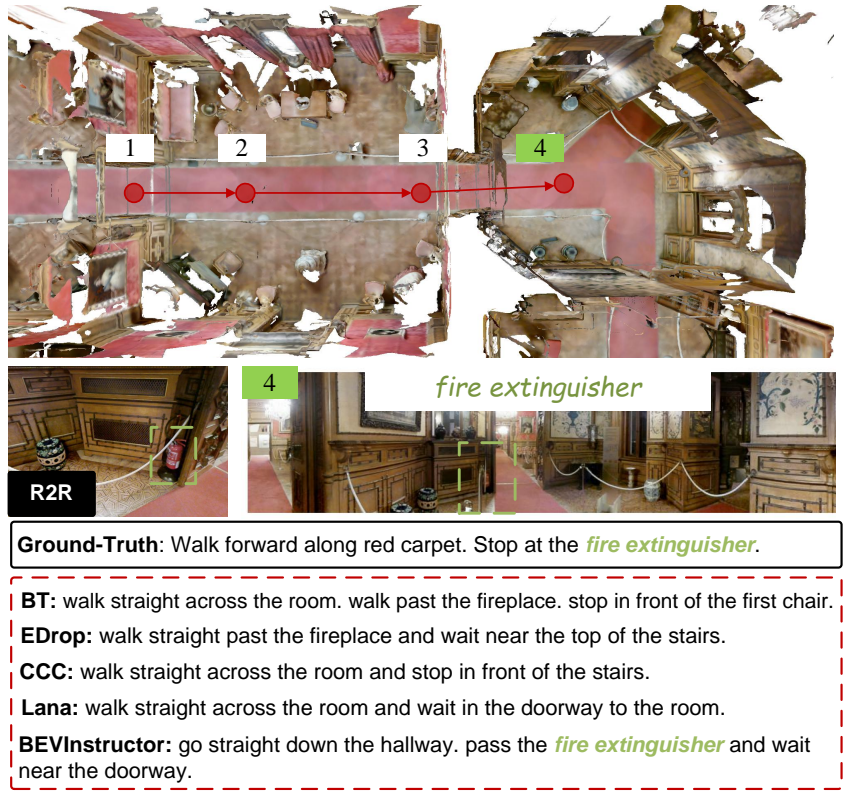}
	\captionsetup{font=small}
	\caption{\small Comparison results among Ground-Truth, BT-speaker~\cite{fried2018speaker}, EDrop-speaker~\cite{tan2019learning}, CCC-speaker~\cite{wang2022counterfactual}, Lana~\cite{wang2023lana}, and \textsc{BEVInstructor} for instruction generation on R2R~\cite{anderson2018vision} val unseen split. See \S\ref{sec:supplequan} for more details.}
	\label{fig:qualitative2}
\end{figure*}

\noindent\textbf{User Study.} Due to the inherent limitations of evaluation metrics, the current metrics cannot fully reflect the performance of generated instructions~\cite{zhao2021evaluation}. To more comprehensively reflect its performance, we conduct a set of human evaluation experiments. Specifically, 23 college students are invited to evaluate 100 instructions in total generated by various algorithms, including \textsc{BEVInstructor}, BT-speaker, EDrop-speaker, CCC-speaker, and Lana. They score each instruction based on its match with the navigation path using a scale from 0 to 5. The test paths for user study are sampled from REVERIE \texttt{val unseen}. As a result, \textsc{BEVInstructor}, with a score of $3.64$, outperforms the other models, \ie, BT-speaker $2.97$, EDrop-speaker $3.11$, CCC-speaker $3.24$, and Lana $3.43$. 

\noindent\textbf{More Examples.} In Fig.\!~\ref{fig:qualitative2}, \textsc{BEVInstructor} successfully captures the crucial landmarks, \eg, \textit{fire extinguisher}, based on the perspective-BEV module, while other algorithms fail to provide more detailed information in their instructions.

\section{Discussion}
\label{sec:discussion}
\noindent\textbf{Limitations.} \textsc{BEVInstructor} outperforms existing methods across all datasets, \ie, R2R~\cite{anderson2018vision}, REVERIE~\cite{reverie}, and UrbanWalk~\cite{huang2022assister}. Despite notable progress, in comparison to human-annotated instructions, there exists considerable space for enhancing the diversity and accuracy of the instructions. In open-vocabulary settings, \textsc{BEVInstructor} continues to necessitate human intervention for the purpose of filtering and correction. Moreover, \textsc{BEVInstructor} currently does not incorporate safety factors, \eg, warnings of dangerous areas, which are crucial for application in real-world scenarios.

\noindent\textbf{Social Impact.} The proposed framework for navigation instruction generation, incorporating MLLMs and BEV features, presents a pioneering contribution with substantial implications for social impact. Our approach not only achieves an impressive improvement in the performance, but also has stronger interpretability through outputting landmarks in the process of refinement. This approach can significantly enhance the trust between humans and agents during navigation, aligning more closely with human cognitive methods.

\noindent\textbf{Future Work.} \textsc{BEVInstructor} integrates perspective features and BEV features into a unified representation. Given that the BEV coordinate is consistent with the 3D coordinate, BEV naturally supports multi-sensor fusion~\cite{li2023delving,ma2022vision,NEURIPS2022_43d2b7fb,10160968}. Future developments for \textsc{BEVInstructor} aim to expand this framework by incorporating the 3D world into the current MLLM via multi-sensor features, \eg, LiDAR. This advancement will not only contribute to the robustness and versatility of \textsc{BEVInstructor} but also elevate its efficacy in real-world scenarios. Furthermore, recognizing the importance of safety, future enhancements will focus on embedding navigational and environmental safety measures into the model.

\end{document}